\documentclass[letterpaper]{article}
\usepackage[letterpaper, margin=1in]{geometry}
\usepackage{times}  
\usepackage{helvet} 
\usepackage{courier} 
\usepackage[hyphens]{url} 
\usepackage{graphicx}
\usepackage[numbers,sort]{natbib}
\usepackage{caption} 
\DeclareCaptionStyle{ruled}{labelfont=normalfont,labelsep=colon,strut=off} 
\usepackage{algorithm}
\usepackage{algpseudocode}
\usepackage{bm}
\usepackage{bbm}
\usepackage{amsmath,amssymb,amsfonts}
\usepackage{amsthm}

\usepackage{subfig}
\usepackage{multirow}
\usepackage{xcolor}
\newcommand{\myparatight}[1]{\smallskip\noindent{\bf {#1}:}~}
\newcommand{\argmax}{\operatornamewithlimits{argmax}}

\newcounter{equal}

\title{FaceGuard: Proactive Deepfake Detection}
\author {
    Yuankun Yang\textsuperscript{1}\setcounter{equal}{\value{footnote}}\thanks{The first two authors made equal contributions. They performed this research when they were remote interns in Gong's group.},
    Chenyue Liang\textsuperscript{2}\setcounter{footnote}{\value{equal}}\footnotemark,
    Hongyu He\textsuperscript{3},
	 Xiaoyu Cao\textsuperscript{3},
	 Neil Zhenqiang Gong\textsuperscript{3}
}

\date{%
	\textsuperscript{1}Fudan University,  17307110068@fudan.edu.cn\\%
	\textsuperscript{2}Chinese Academy of Sciences,  lllcy\_cheryl@outlook.com\\%
	\textsuperscript{3}Duke University,  \{hongyu.he, xiaoyu.cao, neil.gong\}@duke.edu
}

\begin{document}
\maketitle


\begin{abstract}
Existing deepfake-detection methods focus on \emph{passive} detection, i.e., they detect fake face images via exploiting the artifacts produced during deepfake manipulation. A key limitation of passive detection is that it cannot detect fake faces that are generated by new deepfake generation methods. 
In this work, we propose \emph{FaceGuard}, a \emph{proactive} deepfake-detection framework. FaceGuard embeds a watermark into a real face image before it is published on social media. Given a face image that claims to be an individual (e.g., Nicolas Cage), FaceGuard extracts a watermark from it and predicts the face image to be fake if the extracted watermark does not match well with the individual's ground truth one. A key component of FaceGuard is a new deep-learning-based watermarking method, which is 1) robust to normal image post-processing such as JPEG compression, Gaussian blurring, cropping, and resizing, but 2) fragile to deepfake manipulation. Our evaluation on multiple datasets shows that  FaceGuard can detect deepfakes accurately and outperforms existing methods. 
\end{abstract}


\section{Introduction}
As deep learning becomes more and more powerful, deep learning based \emph{deepfake generation methods} can produce more and more realistic-looking deepfakes \cite{karras2017progressive,karras2019style,choi2018stargan,park2019semantic,karras2020analyzing,zhu2017unpaired,thies2016face2face,thies2019deferred,nirkin2019fsgan,zakharov2019few}. In this work, we focus on fake faces because faces are key ingredients in human communications. Moreover, we focus on \emph{manipulated} fake faces, in which a deepfake generation method replaces a target face as a source face (known as \emph{face replacement}) or changes the facial expressions of a target face as those of a source face (known as \emph{face reenactment}). For instance, in the well-known Trump-Cage deepfakes example~\cite{trumpcage}, Trump's face (target face) is replaced as Cage's face (source face). Fake faces can be used to assist the propagation of fake news, rumors, and disinformation on social media (e.g., Facebook, Twitter, and Instagram). Therefore, fake faces pose growing concerns to the integrity of online information, highlighting the urgent needs for deepfake detection.

Existing deepfake detection mainly focuses on \emph{passive} detection, which exploits the artifacts in fake faces to detect them after they have been generated.   
Specifically, given a face image, a passive detector extracts various features from it and classifies it to be real or fake based on the features. The features can be manually designed based on some heuristics~\cite{agarwal2019protecting,li2018ictu,matern2019exploiting,yang2019exposing,li2019exposing,frank2020leveraging} or automatically extracted by a deep neural network based feature extractor~\cite{zhou2017two,afchar2018mesonet,roessler2019faceforensicspp,nguyen2019multi,nguyen2019use,cozzolino2017recasting,bayar2016deep,rahmouni2017distinguishing,wang2019cnngenerated,frank2020leveraging}.  Passive detection faces a key limitation \cite{cao2021understanding}, i.e., it cannot detect fake faces that are generated by new deepfake generation methods that were not considered when training the passive detector. As new deepfake generation methods are continuously developed, this limitation poses significant challenges to passive deepfake detection.  

\begin{figure}[!t]
\centering
\includegraphics[width=0.75\textwidth]{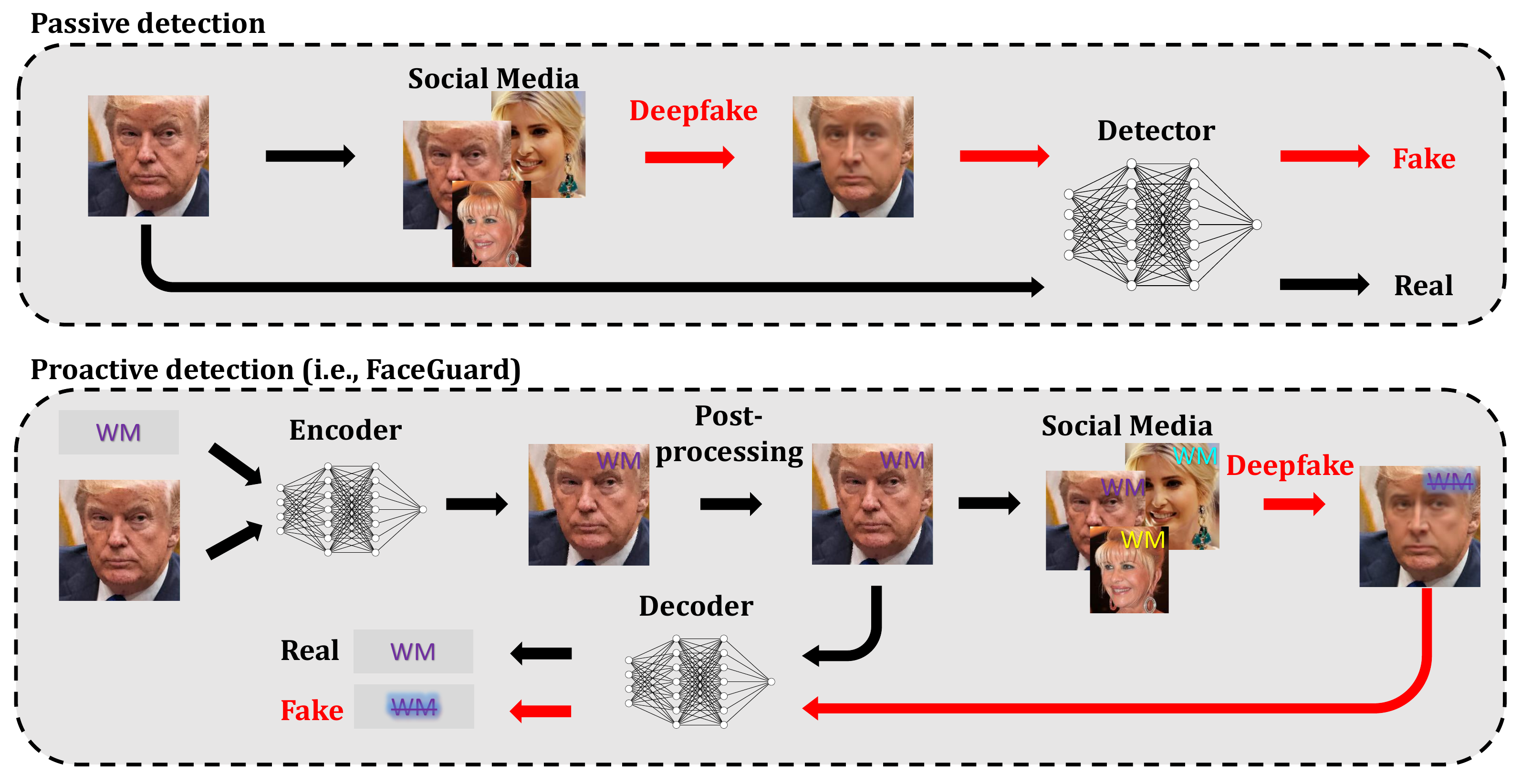}
\vspace{-1mm}
\caption{FaceGuard v.s. passive detection.}
\label{fig:universe} 
\vspace{-4mm}
\end{figure}

\myparatight{Our work} In this work, we propose \emph{FaceGuard}, a \emph{proactive} deepfake-detection framework. FaceGuard addresses the limitation of passive detection via proactively embedding watermarks into real face images before they are manipulated by deepfake generation methods. Figure~\ref{fig:universe} illustrates the difference between passive detection and FaceGuard. 
 Specifically, before posting an individual's real face image on social media, FaceGuard embeds a watermark (i.e., a binary vector in our work) into it. The watermark is human imperceptible, i.e., a face image and its watermarked version look visually the same to human eyes. For instance, the watermark can be embedded into an individual's face image using the individual's smartphone.  
 Suppose a face image is claimed to be an individual, e.g., the manipulated face is claimed to be Nicolas Cage in the Trump-Case deepfakes example. FaceGuard extracts a watermark from the face image and predicts it to be fake if the fraction of matched bits between the extracted binary-vector watermark and the individual's ground truth one is smaller than a threshold. 

The key components of FaceGuard include an \emph{encoder} and a \emph{decoder}. Both the encoder and decoder are neural networks, but they may have different architectures and parameters. The encoder takes a face image and a watermark as input, and outputs a watermarked face image. The decoder takes a (watermarked) face image as input and outputs a binary-vector watermark. FaceGuard aims to achieve two key properties. On the one hand,  a watermarked face image may be post-processed, e.g., via JPEG compression, Gaussian blurring, cropping, and resizing. For instance, during transmission on the Internet, images are often JPEG compressed to reduce their sizes. Our watermarks in FaceGuard should be robust to such post-processing, i.e., our decoder can still extract an individual's watermark from its post-processed watermarked face images. Therefore, FaceGuard can still predict such post-processed watermarked real face images as real.  On the other hand, an attacker downloads a watermarked face image from a social media, uses a deepfake generation method to manipulate it, and tries to re-post the fake face image on the social media. Our watermarks should be fragile to deepfake manipulation, i.e., our decoder cannot extract the individual's watermark from a fake face image claiming as the individual. Therefore, the social media can detect the fake face images using FaceGuard. 

 To achieve the two properties above, we jointly train the encoder and decoder using a set of real face images. In particular, we formulate a loss function, which consists of a \emph{reconstruction loss} of the watermark as well as a \emph{pixel loss} and an \emph{adversarial loss} that capture the perceptual difference between a real face image and its watermarked version. We learn an encoder and decoder via minimizing the loss function using stochastic gradient descent.  Moreover, we add a \emph{post-processing module} between the encoder and decoder, which applies post-processing operations of interest to the watermarked images produced by the encoder before feeding them into the decoder. However, the post-processing module introduces two key challenges to training. The first challenge is that, JPEG compression, a popular post-processing operation, is non-differentiable, making it impossible to backpropagate  gradients from decoder to encoder. The second challenge is how to schedule different post-processing operations for each mini-batch of training images, given multiple post-processing operations. 

To address the first challenge, we propose to approximate the quantization/dequantization components of JPEG, which enables us to backpropagate meaningful gradients during training. To address the second challenge, we propose a reinforcement learning based scheduler to determine which post-processing operation should be applied for each mini-batch of training images. 

We empirically evaluate FaceGuard using multiple datasets, including  FaceForensics~\cite{roessler2019faceforensicspp}, Facebook Deepfake Detection Challenge~\cite{DFDC2020},  and a Trump-Cage dataset, as well as three deepfake generation methods (two face replacement methods and one face reenactment method). Our results show that FaceGuard can accurately detect fake faces generated by different deepfake generation methods; FaceGuard outperforms passive detection; our watermarking method in FaceGuard outperforms existing ones; and FaceGuard is robust against adaptive deepfakes in which an attacker trains its own encoder/decoder to embed watermarks into its deepfakes.  

In summary, our contributions are as follows:
\begin{itemize}
    \item We propose FaceGuard, a proactive deepfake-detection framework. 
    \item We propose a method to make JPEG compression differentiable and a reinforcement learning based method to schedule post-processing operations when jointly training the encoder and decoder in FaceGuard. 
    \item We extensively evaluate FaceGuard and compare it with state-of-the-art methods.
\end{itemize}


\section{Related Work}
\subsection{Deepfake Generation}
Fake face images can be generated via \emph{face synthesizing} \cite{karras2017progressive,karras2019style,choi2018stargan,park2019semantic,karras2020analyzing,zhu2017unpaired} or \emph{face manipulation} \cite{thies2016face2face,thies2019deferred,nirkin2019fsgan,faceswap1,zakharov2019few}. Face synthesizing creates new faces that do not belong to any individual in the world, while face manipulation manipulates the face images of existing individuals. Generally speaking, face manipulation may result in more severe consequences than face synthesizing, as it may target individuals with high social impact, e.g.,  President of United States and celebrities. Therefore, we focus on face manipulation in this work. 

Face manipulation can be further categorized into \emph{face replacement} \cite{faceswap1,zakharov2019few} and \emph{face reenactment} \cite{thies2016face2face,thies2019deferred,nirkin2019fsgan}. In face replacement, a target face is replaced as a source face and the generated fake face claims to be the source. For instance, given face images of a target individual and a source individual, FaceSwap \cite{faceswap1} trains an autoencoder for each individual, where the encoders of the two autoencoders are shared.  When generating fake faces, FaceSwap first encodes a target face image using the shared encoder and then decodes it using the source's decoder, which transforms the target face into the source face. In face reenactment, the expression, hair color, and/or other properties of the target face are changed as those of a source face. For instance, ICface \cite{tripathy+kannala+rahtu} trains GANs in a self-supervised manner to utilize human interpretable control signals on emotions and head poses. Face replacement changes the identity of the target face to that of the source, while face reenactment preserves the target's face identity.

\subsection{Deepfake Detection}
Existing deepfake-detection methods are mainly \emph{passive}.  
These passive deepfake-detection methods fall into two categories, i.e., \emph{heuristics-based methods} \cite{agarwal2019protecting,li2018ictu,matern2019exploiting,yang2019exposing,li2019exposing,frank2020leveraging} and \emph{deep-learning-based methods} \cite{zhou2017two,afchar2018mesonet,roessler2019faceforensicspp,nguyen2019multi,nguyen2019use,cozzolino2017recasting,bayar2016deep,rahmouni2017distinguishing,wang2019cnngenerated,frank2020leveraging,subramani2020learning,wangfakespotter,zhangdetecting,hudynamic,he2021beyond}. Heuristic-based methods leverage some heuristic features to detect fake faces. For instance, Li et al. \cite{li2018ictu} detected fake faces by analyzing  eye-blinking of the person in a video.  Deep-learning-based methods train binary deep neural network (DNN) classifiers to perform detection, i.e., a binary DNN classifier takes a face image as input and outputs fake or real.  Deep-learning-based methods outperform heuristics-based methods. 
However, a recent study \cite{cao2021understanding} shows that an attacker can evade passive detection via various strategies, e.g., an attacker can use a new deepfake generation method to generate fake faces that will be misclassified as real by a passive detector. 

Recently, Wang et al. proposed FakeTagger \cite{wang2021faketagger}, which aims to track image reuse on social media. The idea is to embed tags to real face images. The tags are robust against both normal post-processing and deepfake manipulations of the tagged face images. FakeTagger predicts a face image as fake if the tag can be extracted from it. However, FakeTagger falsely predicts normally post-processed versions of real face images as fake because the tags are robust against normal post-processing. For instance, if a real face image is downloaded from a social media, JPEG compressed, and re-uploaded to the social media, it will be falsely flagged as fake by FakeTagger. In contrast, our FaceGuard leverages semi-fragile watermarks, which are robust against normal post-processing but are fragile to deepfake manipulations. Therefore, FaceGuard can  distinguish  fake faces from  normally post-processed real ones.

\subsection{Watermarks}

Watermarking \cite{tirkel1993electronic} was originally proposed for copyright protection. For instance,  a watermark can be embedded into an image as a signature of ownership, which can be extracted later for ownership verification. Conventional watermarks include distortions in both \emph{spatial domain} \cite{tirkel1993electronic,van1994digital} and \emph{transform domain}, such as discrete cosine transform \cite{barni1998dct,chu2003dct}, discrete wavelet transform \cite{ganic2004robust,al2007combined}, and discrete Fourier transform \cite{deguillaume1999robust,ruanaidh1996phase}. The key limitation of conventional watermarks is that they are not robust enough to common image post-processing \cite{zhu2018hidden}. 

To address such limitation, recent works \cite{zhu2018hidden,zhang2020udh,ahmadi2020redmark,liu2019novel,luo2020distortion} proposed deep-learning-based watermarks. Specifically, in these methods, an \emph{encoder} takes an image and a watermark as input and outputs a watermarked image, while a \emph{decoder} takes a (watermarked) image as input and outputs a watermark. Moreover, given a set of images, they aim to train the encoder and decoder such that the decoder can still extract the correct watermark from a watermarked image even if it has normal post-processing such as JPEG compression. 

For instance, HiDDeN \cite{zhu2018hidden} proposed a post-processing module between the encoder and decoder during training, where the post-processing module applies post-processing to the watermarked images. In particular, for each mini-batch  during training, HiDDeN randomly picks one post-processing from a set of predefined ones uniformly at random and applies it to the watermarked images in the mini-batch.  A key challenge of such post-processing module based training is that JPEG compression is not differentiable, which means that the gradients cannot be  backpropagated from the decoder to encoder during training. To address this challenge,  HiDDeN proposed to remove the non-differentiable quantization/dequantization steps of JPEG compression and simply set high-frequency coefficients in DCT as zero. However, as we will show in experiments, HiDDeN is not robust to JPEG compression. 


\section{Training Encoder and Decoder in FaceGuard}

\begin{figure}[!t]
\centering
\includegraphics[width=0.75\textwidth]{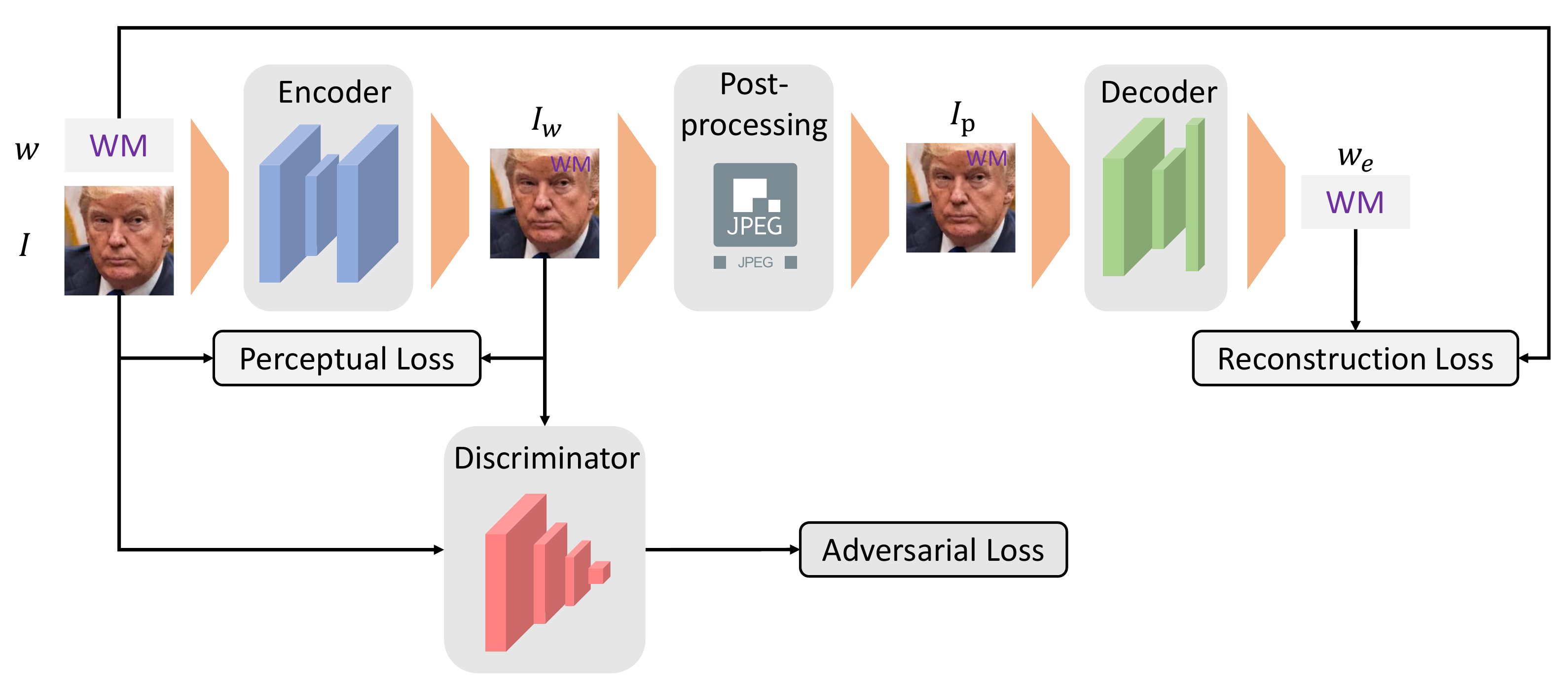}
\vspace{-1mm}
\caption{Overview of jointly training the encoder and decoder in FaceGuard.}
\label{fig:faceguard} 
\vspace{-4mm}
\end{figure}

Figure \ref{fig:faceguard} shows the end-to-end training workflow of our encoder and decoder. Specifically, we use binary vectors as our watermarks. Given a real face image $I$ and a binary-vector watermark $w$, the encoder outputs a watermarked face image $I_w$. To achieve robustness against normal post-processing, we place a post-processing module in between the encoder and the decoder. This module performs a normal post-processing operation on a watermarked image and produces a post-processed watermarked image $I_p$. During our training, the module switches between different normal post-processing operations in different mini-batches. In particular, for each mini-batch of the training images, we select one normal post-processing operation and apply it in the module.
The decoder extracts a binary-vector watermark $w_e$ from a post-processed watermarked image $I_p$. 

\myparatight{Formulating an optimization problem} We formulate jointly learning the encoder and decoder as an optimization problem. 
First, the extracted watermark $w_e$ produced by the decoder should be similar to the true watermark $w$. Therefore, we formulate a \emph{reconstruction loss} $\Vert w_e - w \Vert_2^2$. Second,  we aim to produce human imperceptible watermarks, i.e., an image $I$ and its watermarked version $I_w$ should be visually similar. To achieve this goal, we formulate a \emph{pixel loss} and an \emph{adversarial loss}. In particular, we use $\Vert I_w - I \Vert_2^2$ as a pixel loss, which measures the  $\ell_2$-norm of the difference between an image and its watermarked version. Moreover, inspired by \cite{zhu2018hidden}, we use a discriminator  to distinguish between real images and their watermarked versions. Given an image $I$, the discriminator outputs $D(I)$, i.e., the probability that $I$ is a non-watermarked image. We aim to train the encoder such that its output $I_w$ has high probability $D(I_w)$ predicted by the discriminator. Therefore, we use $\log(1-D(I_w))$ as an adversarial loss. To summarize, given a discriminator $D$, we formulate a loss function $\mathcal{L} = \Vert w_e - w \Vert_2^2 + \lambda_1\Vert I_w - I \Vert_2^2 + \lambda_2 \log(1-D(I_w))$ for the encoder and decoder, where $\lambda_1$ and $\lambda_2$ are used to balance the pixel loss and adversarial loss. In our framework, we also jointly learn the discriminator.  Formally, we jointly learn the encoder, decoder, and discriminator via solving the following optimization problem:

\begin{align}
\label{optproblem}
    &\underset{\theta,\phi,\psi}{\text{min }} \mathbb{E}_{I,w}[\mathcal{L}]+\mathbb{E}_{I,w}[\log(1-D(I))+\log(D(I_w))],
\end{align}
where $\theta, \phi$, and $\psi$ are the parameters of the encoder, the decoder, and the discriminator, respectively; and the expectation is taken with respect to the distribution of real face image $I$ and  distribution of watermark $w$. In experiments, we use a  set of real face images and generate random binary-vector watermarks in each mini-batch of training images to calculate the expectation.    

\myparatight{Solving the optimization problem} We alternatively optimize the encoder/decoder and the discriminator. Specifically, given the current discriminator, we update the encoder and decoder via solving Equation~(\ref{optproblem}) using stochastic gradient descent; and given the current encoder and decoder, we update the discriminator via solving Equation~(\ref{optproblem}) using stochastic gradient descent. However, there are two key challenges when updating the encoder and decoder for a given discriminator. First, although the post-processing module does not have trainable parameters,  it needs to be differentiable for backpropagating  gradients from the decoder to encoder. Despite most post-processing operations satisfy the differentiability requirement, JPEG, one of the most common and important post-processing operations, is not differentiable. Second, for each mini-batch of the training images, we need to select a post-processing operation to apply. A naive approach is to  pick a post-processing operation among a set of predefined ones uniformly at random, as used by HiDDeN. However, this naive scheduling leads to sub-optimal performance as shown in our experiments. 

To address the first challenge, we propose an approximation of JPEG compression that makes backpropagation possible. Moreover, we leverage reinforcement learning to schedule the different post-processing operations to address the second challenge. Next, we discuss the details. 

\begin{figure}[!t]
\centering
\hfill
\subfloat[Original JPEG]{\includegraphics[width=0.45\textwidth]{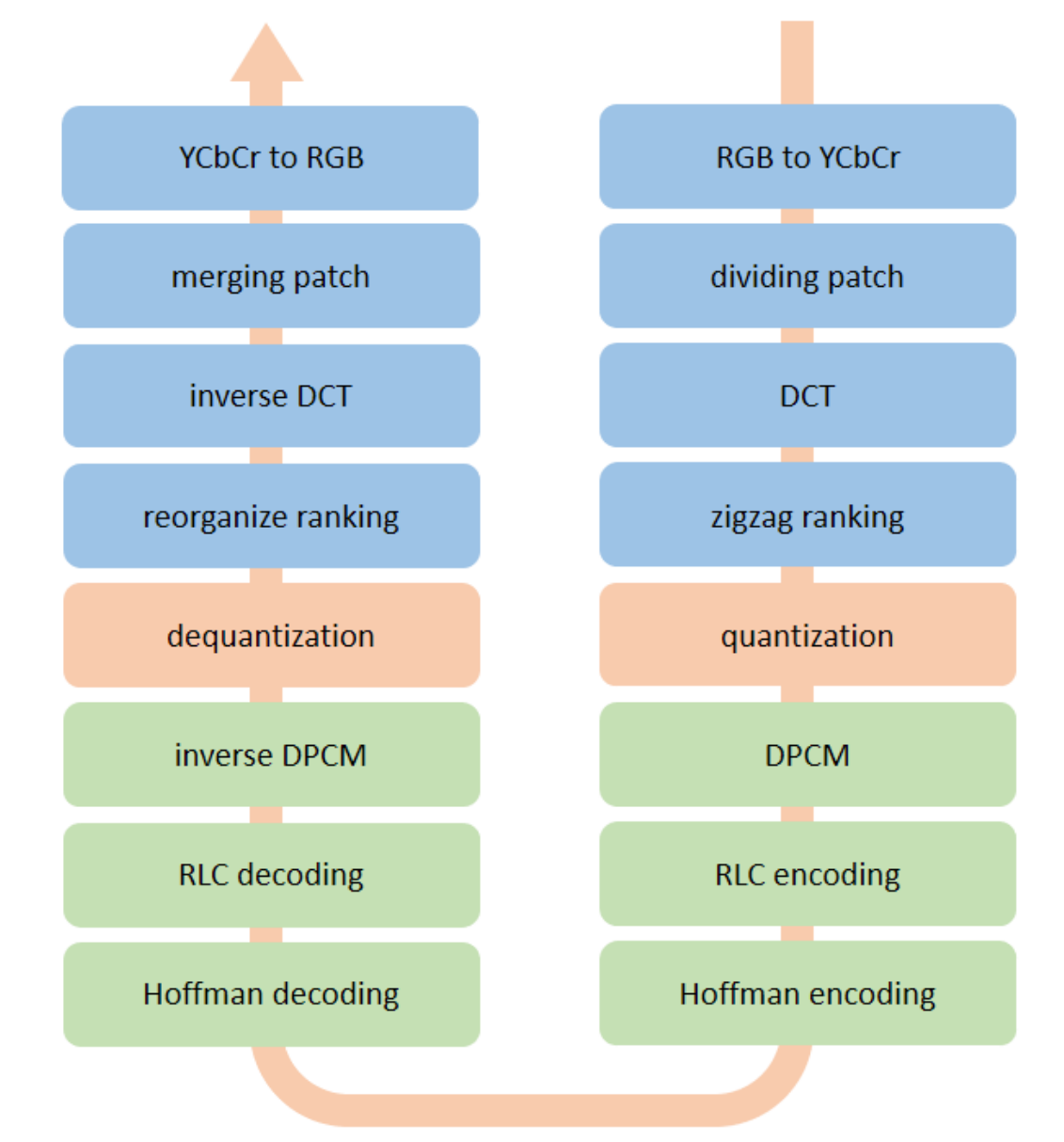}\label{fig:jpeg}}\hfill
\subfloat[Our simplified JPEG]{\includegraphics[width=0.45\textwidth]{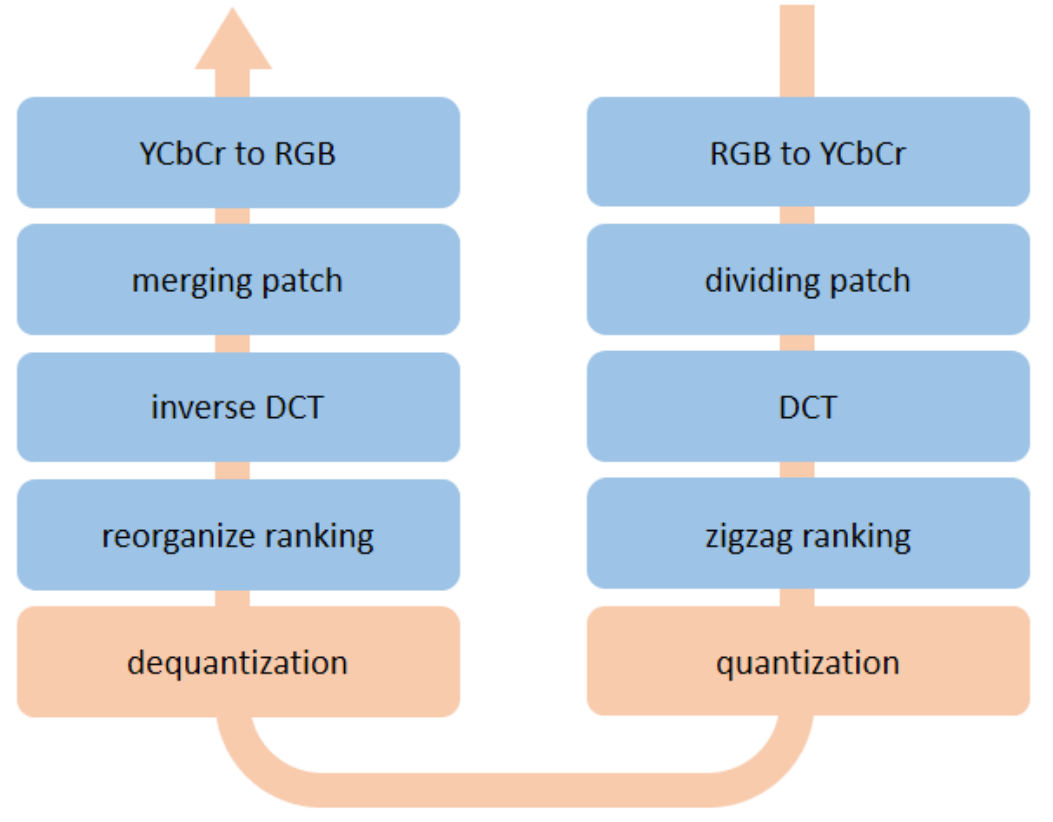}\label{fig:simp_jpeg}} \hfill
\caption{Illustration of the original JPEG compression procedure and our simplified JPEG compression procedure.}
\end{figure}

\subsection{Approximating JPEG Compression for Backprop}
Figure \ref{fig:jpeg} illustrates how JPEG compression works. Most of the steps are differentiable, except the quantization/dequantization ones. 
For simplicity, we omit the steps between quantization and dequantization  as they are invertible operations, whose combination is equivalent to the identity mapping. Our simplified JPEG compression workflow is shown in Figure \ref{fig:simp_jpeg}.

In the quantization step, the input is element-wisely divided by a quantization matrix and rounded to the nearest integers.  In the dequantization step, the rounded values are multiplied by the same quantization matrix to reconstruct the initial input. Specifically, let $x$ denote the input to the quantization step and $g(x)$ denote the output of the dequantization step. Formally, $g(x)$ is defined as follows in JPEG:
\begin{align}
     g(x)=\left\{
\begin{aligned}[rl]
&x-(x\%1),   &x\%1<0.5\\
&x + 1 - (x\%1), &x\%1 \geq 0.5,
\end{aligned}
\right.
\label{eq:fx}
\end{align}
where $x\%1$ represents the decimal part of $x$. Due to the rounding operation in quantization, different inputs to the quantization step may lead to the same output after dequantization. Moreover, the rounding operation makes the gradient through the  quantization-dequantization operations meaningless. Specifically, the gradient $\nabla_x g(x)$ is zero everywhere except when the decimal part of $x$ is 0.5, where $g(x)$ is non-differentiable. Formally, we have:
\begin{align}
     \nabla_x g(x)=\left\{
\begin{aligned}[rl]
&0,   &x\%1\neq 0.5 \\
&none, &x\%1 = 0.5, 
\end{aligned}
\right.
\label{eq:nabla_fx}
\end{align}

To address the aforementioned challenge, we approximate the gradient of the quantization-dequantization steps by explicitly characterizing the changes in $x$. In particular, we  approximate  the quantization-dequantization steps as a linear transformation $g(x)=kx$, where $k$ is as follows:
\begin{align}
k = \left\{
\begin{aligned}[rl]
&\frac{x-x\%1}{x},   &x\%1<0.5 \land x\neq 0\\
&\frac{x+1-x\%1}{x}, &x\%1 \geq 0.5\\
&0, &x=0.
\end{aligned}
\right.
\end{align}
Here, $k$ is the ratio between the output and the input of the quantization-dequantization steps, which quantifies the change to $x$ introduced by these steps. Therefore, we use it to approximately represent the gradient. Specifically, in the forward pass, we calculate the value of $g(x)$ following Equation (\ref{eq:fx}). However, in the backward pass, instead of using Equation \ref{eq:nabla_fx}, we treat $k$ as a constant and consider it as our gradient, i.e., we have $\nabla_x g(x)=k$. Our approximation makes it possible to obtain meaningful end-to-end gradients in our training workflow, which enables us to learn the encoder and decoder via stochastic gradient descent. 

\subsection{Scheduling Post-processing via Reinforcement Learning}
Given multiple commonly used post-processing operations, the post-processing module selects one  for each mini-batch of training images. The \emph{random scheduler}, which picks a post-processing operation uniformly at random, achieves sub-optimal performance because the difficulty of fitting different post-processing operations varies. For instance, assume we have two post-processing operations $p_1$ and $p_2$. $p_1$ is easy to fit, while $p_2$ is hard. A random scheduler essentially switches between $p_1$ and $p_2$ with equal probability. At some point during training, the encoder/decoder fits $p_1$ well, while still underfits $p_2$ due to the different difficulty levels. Beyond this point, ideally the scheduler should focus on $p_2$. However, a random scheduler will still vacillate between the two, leading to overfit $p_1$ and/or underfit $p_2$. 

Therefore, we design an advanced scheduler using reinforcement learning, which learns to adaptively switch among different post-processing operations. Specifically, 
we consider the choices of different post-processing operations as our states, and the switches between different choices as the set of actions. Assuming there are $N$ post-processing operations. We denote the set of states as $\mathcal{S}=\{s^{(i)}|i=1,2,\cdots,N\}$ and the set of actions as $\mathcal{A}=\{a^{(i)}|i=1,2,\cdots,N\}$, where $a^{(i)}$ means the state will switch to $s^{(i)}$ for the next mini-batch. For an input image, we define its \emph{bitwise accuracy} as the fraction of bits in the extracted binary-vector watermark $w_e$ that match those in $w$. For a mini-batch $B$, we define its average bitwise accuracy $f_t$  as the mean of the bitwise accuracy of  images in the mini-batch. We further design our reward function as follows:
\begin{align}
    r(s_t, a_t) = \beta[f_t-h_{s_{t+1}}] + f_t + b_{s_{t+1}}.
    \label{eq:reward}
\end{align}
In the reward function, $\beta$ is a constant weight and $s_{t+1}$ is the state for the $(t+1)$-th mini-batch, which depends on the action $a_{t}$. $\mathbf{h}$ is an $N$-dimensional vector that records the previous bitwise accuracy for each state. $\mathbf{b}$ is a constant $N$-dimensional vector that represents a bias for each state in the reward function based on the domain knowledge. $h_{s_{t+1}}$ and $b_{s_{t+1}}$ are the $s_{t+1}$-th entry in $\mathbf{h}$ and $\mathbf{b}$, respectively.

We train our scheduler following  Q-learning \cite{watkins1992q} in an online learning manner, i.e., we keep updating our scheduler at the same time we train our encoder and decoder. Specifically, we keep a $N\times N$ matrix of Q-values called Q-table. Each row in the matrix represents a state, while each column represents an action. The Q-value at location $(i,j)$ indicates the benefit of taking action $a^{(j)}$ at state $s^{(i)}$. At the beginning, we initialize the Q-table $Q$ with arbitrary values (e.g., all zeros), initialize the vector $\mathbf{h}$ as all zeros, and start with a random state. Later, for each mini-batch $B_t$, we decide which post-processing to apply to $B_t$ based on the Q-table. Considering the exploitation-exploration trade-off, with probability $1-\epsilon_t$, we select the action with the largest Q-value at state $s_t$. With probability $\epsilon_t$, we randomly pick an action to take. We set $\epsilon_t$ to be large at the beginning and decay it gradually. We perform the selected post-processing in between the encoder and the decoder, and compute the average bitwise accuracy $f_t$. With $f_t$, we update the history bitwise accuracy vector $\mathbf{h}$ and calculate the reward function $r_t$ following Equation (\ref{eq:reward}). Moreover, we use $r_t$ to update the Q-table following the Q-learning update rule, i.e., we have:
\begin{align}
    Q(s_{t},a_{t})\gets Q(s_{t},a_{t})
    +\alpha [r_t+\gamma \max_{a\in\mathcal{A}} Q(s_{t+1},a)- Q(s_{t},a_{t})],
    \label{eq:update_q}
\end{align}
where $\alpha$ is the learning rate and $\gamma$ is a constant discount factor.
Our scheduler is summarized in Algorithm \ref{alg:rl}. 

\begin{algorithm}[t]
	\caption{Reinforcement-Learning-based Scheduler}\label{alg:rl}
	\begin{algorithmic}[1]
		\renewcommand{\algorithmicrequire}{\textbf{Input:}}
		\renewcommand{\algorithmicensure}{\textbf{Output:}}
		\Require  $\mathcal{S}$, $\mathcal{A}$, $r$, $\mathbf{h}$, $\mathbf{b}$, $\epsilon_t$, $s_t$, $B_t$, and $Q$.
		\Ensure  Selected post-processing operation for $B_t$.
		\State $u \gets \text{Uniform}(0,1)$
		\If{$u < 1-\epsilon_t$} \hfill// exploitation
    		\State $a_t \gets \argmax_{a\in\mathcal{A}} Q(s_{t},a)$
		\Else \hfill// exploration
		\State $a_t \gets $ a random action $a\in\mathcal{A}$
		\EndIf
		\State Calculate average bitwise accuracy $f_t$
		\State Update the Q-table  based on Equation (\ref{eq:update_q})
		\State Decay $\epsilon_t$
		\State $h_{s_{t+1}} \gets f_t$ \hfill //update $\mathbf{h}$
		\\\Return $a_t$
	\end{algorithmic} 
\end{algorithm}


\section{Experiments}
\subsection{Experimental Setup}

\myparatight{Datasets} We use face images from three popular datasets, i.e., FaceForensics++ \cite{roessler2019faceforensicspp}, Facebook Deepfake Detection Challenge (DFDC) \cite{DFDC2020}, and Trump-Cage \cite{trumpcagedata}.

{\bf FaceForensics++.} FaceForeinsics++ consists of YouTube videos that have faces in them. We randomly pick two videos and extract face images from them. We consider one of them as the target and the other as the source. For each video, we split it into frames and extract face images from the frames using the publicly available package Dlib~\cite{dlib09}. We follow \cite{roessler2019faceforensicspp} to enlarge the face regions around the located center  by 1.3$\times$.  We also resize the face images to $299\times299$ pixels. In total, we have 460 target face images and 693 source face images. 

{\bf DFDC.} DFDC contains videos made by paid actors.  
Like FaceForeinsics++, we randomly select two videos from DFDC, each containing face images of one individual; we extract face images from them; and we enlarge the face regions and resize  the face images to $299\times299$ pixels. In total, we have 300 face images of the target and 300 face images of the source. 

{\bf Trump-Cage.} The Trump-Cage dataset contains face images of  Donald Trump and  Nicolas Cage, which are crawled from the Internet. Specifically, there are 376 face images of Trump (target) and 318 face images of Cage (source). 

For each of the three datasets, we use 60\% of face images as the training dataset, 20\% as the validation dataset, and the remaining 20\% as the test dataset. 

\myparatight{Deepfake generation methods}
We consider three deepfake generation methods, i.e., \emph{FaceSwap1} \cite{faceswap1}, \emph{FaceSwap2} \cite{faceswap2}, and \emph{ICface}~\cite{tripathy+kannala+rahtu}. Both FaceSwap1 and FaceSwap2 are face replacement methods, while ICface is a face reenactment method. 

For FaceSwap1 and FaceSwap2, we use the target faces to train the target decoder, the source faces to train the source decoder, and both target and source faces to train the shared encoder. We generate fake faces by feeding the target faces into the encoder and then decoding them using the source decoder. For ICface, we randomly select one target face from each dataset. Then for each source face in the dataset, we generate one fake face. Specifically, we extract the expression from the source face and use it to reenact the target face with the pre-trained model published by the authors.

 \begin{figure}[!t]
 \subfloat[JPEG]{\includegraphics[width=0.25\textwidth]{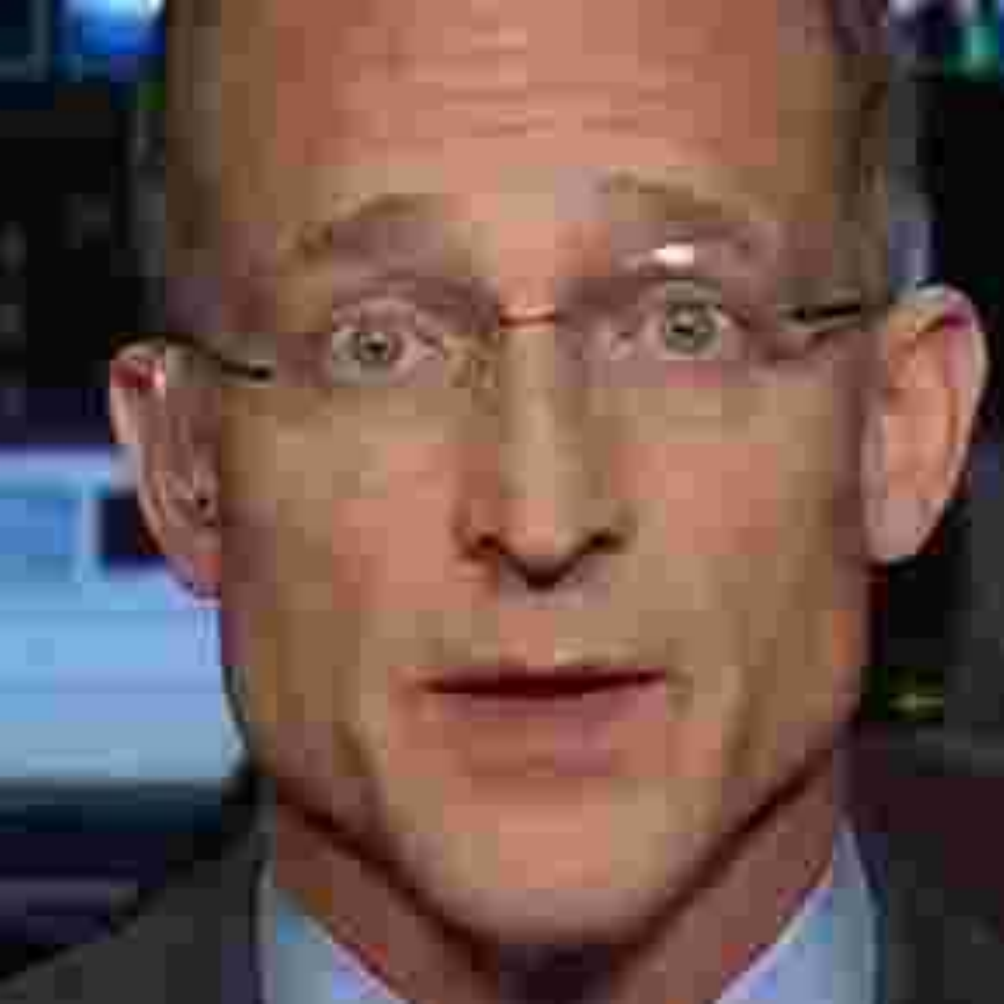}}\hfill
 \subfloat[Gaussian Blur]{\includegraphics[width=0.25\textwidth]{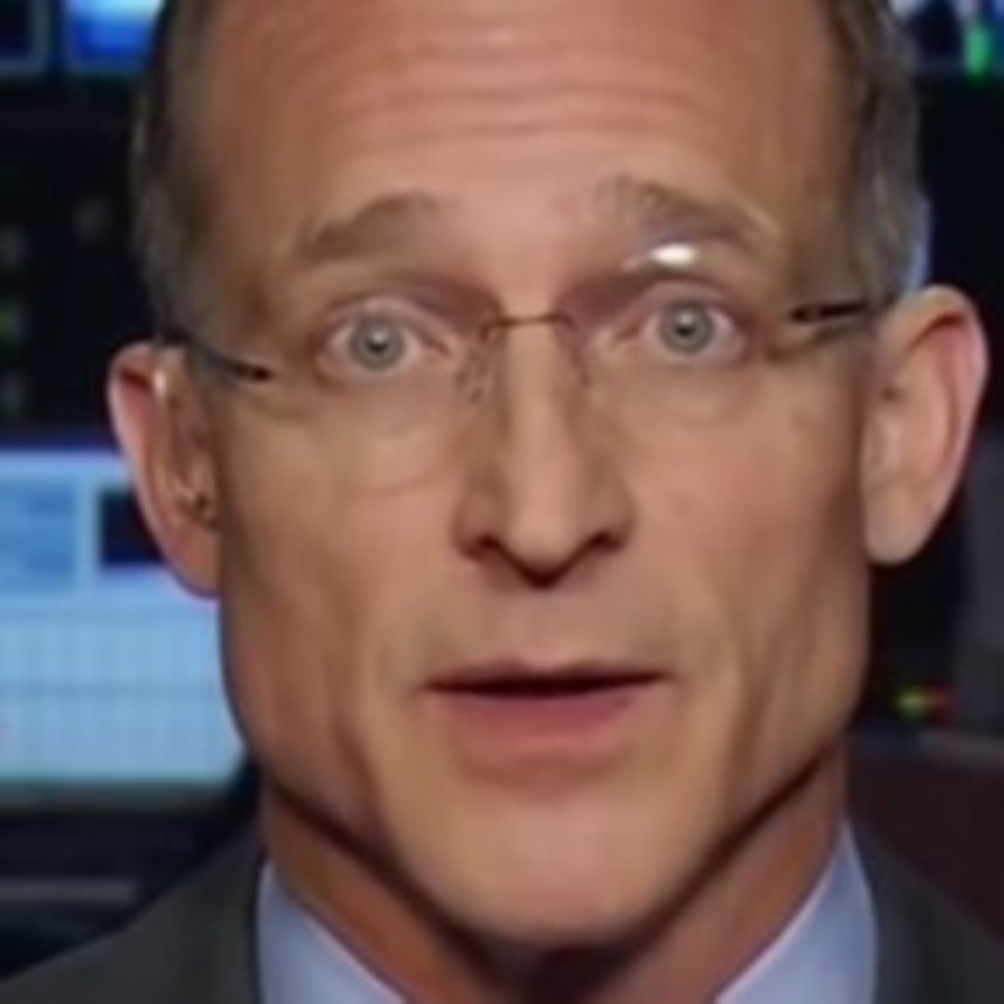}}\hfill
  \subfloat[Crop]{\includegraphics[width=0.15\textwidth]{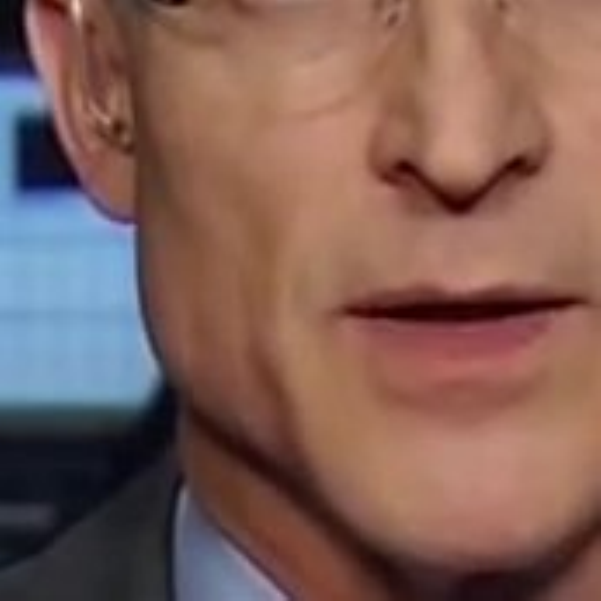}}\hfill
 \subfloat[Resize]{\includegraphics[width=0.15\textwidth]{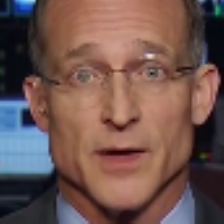} }
\caption{Examples of JPEG compression with quality factor 10, Gaussian blurring with variance 1.0, cropping with size 0.6, and resizing with ratio 0.6.}
\label{pp}
\vspace{-3mm}
 \end{figure}

\myparatight{Post-processing operations} In our experiments, we consider 4 commonly used post-processing operations, i.e., JPEG, Gaussian blurring, cropping, and resizing. For JPEG, we consider quality factors ranging from 10 to 100 with a step size 10. For Gaussian blurring,  we consider  variances ranging from 0 to 1.0 with a step size 0.1. For cropping, we consider crop sizes ranging from 0.6 to 1.0 with a step size 0.1. For resizing, we consider resizing ratios ranging from 0.3 to 1.0 with a step size 0.1. We do not consider crop sizes smaller than 0.6 and resizing ratios smaller than 0.3 because we found they degrade the image quality substantially and are not used in practice. Figure~\ref{pp} shows an example of each post-processing operation. 
For each watermarked real face image, we apply these post-processing operations with different parameters. The real face images and their post-processed versions are treated as \emph{negative} examples, while the generated fake faces are treated as \emph{positive} examples. 

\myparatight{Evaluation metrics} We use Accuracy (ACC), False Positive Rate (FPR), and False Negative Rate (FNR) as our evaluation metrics. ACC is the fraction of test positive and negative examples that are correctly classified. We sample a balanced test set, i.e.,  the same number of testing positive examples and negative examples, to calculate ACC to avoid the bias introduced by unbalanced test set. FPR is the fraction of negative examples that are falsely detected as positive (i.e., fake). FNR is the fraction of positive examples that are incorrectly classified as negative (i.e., real). 

\myparatight{Compared methods} We compare FaceGuard with the following methods. 

{\bf Passive detector.} We consider a state-of-the-art passive detector \cite{roessler2019faceforensicspp}, which trains an Xception network \cite{chollet2017xception} as a detector, i.e., replaces the last classification layer of an Xception net to be a binary classifier. Following \cite{roessler2019faceforensicspp}, we initialize the Xception network with the pre-trained ImageNet weights. We first freeze all the weights except the last layer and train for 3 epochs using the training negative examples (including both real face images and their post-processed versions) and training positive examples generated by FaceSwap1. Then, we train the entire network for another 15 epochs. We use the validation dataset to choose the model with the highest validation accuracy during training. We use FaceSwap1 to train the passive detector to show that its effectiveness degrades for deepfakes generated by other methods not considered during training, following~\cite{cao2021understanding}. 

{\bf Conventional watermarking method (FaceGuard-ConvWM).} Watermarking is a key component of FaceGuard. Our FaceGuard uses a deep-learning-based watermarking method. FaceGuard-ConvWM is a variant of FaceGuard, in which we replace our deep-learning-based watermarking method as a conventional one (both encoder and decoder). Specifically, 
we consider a conventional blind watermarking method \cite{rahman2013conv}. In this watermarking method, the watermark is an image. We predict a face claiming as an individual to be fake if the \emph{structural similarity index measure (SSIM)} between the extracted watermark and the individual's ground-truth one is smaller than a threshold.  

{\bf FaceGuard-HiDDeN.} In this variant, we replace our deep-learning-based watermarking method as HiDDeN \cite{zhu2018hidden}  in FaceGuard. We use the implementation from the authors. 

FaceGuard-ConvWM (or FaceGuard-HiDDeN or FaceGuard) predicts a face image to be fake if the SSIM (or \emph{bitwise accuracy}) between the extracted watermark and the corresponding ground truth one is smaller than a threshold.  Bitwise accuracy is the fraction of matched bits between an extracted watermark and the ground-truth one.  We determine the threshold of each method using the negative examples in the validation dataset.  We select a threshold for a method so the method can detect as many fake faces as possible while falsely classifying a small fraction of real faces as fake. Specifically, we choose the largest threshold for a method such that the method achieves at most 1\% FPR on the validation dataset.  

\begin{table}[!t]
    \centering
    \begin{tabular}{|c|c|}
    \hline
      Parameter & Value \\
      \hline
      \# epochs  & 40,000\\
      \hline
      batch size $|B|$  & 30\\
      \hline
      length of watermark $|w|$  & 30\\
      \hline
      initial learning rate $\eta$ & 0.001\\
      \hline
      loss weights $\lambda_1, \lambda_2$ & 0.7, 0.001\\
      \hline
      reward weight $\beta$ & 10\\
      \hline
      bias vector $\mathbf{b}$ 
      [original, JPEG, Gaussian blurring, cropping, resizing] & [-0.001, 0.001, 0, 0, 0]\\
      \hline
      initial exploration probability $\epsilon_0$ & 1.0\\
      \hline
      exploration probability decay & $2.5\times 10^{-4}$/epoch \\
      \hline
      discount factor $\gamma$ & 0.5\\
      \hline
      scheduler learning rate $\alpha$ & 0.2\\
      \hline
    \end{tabular}
    \caption{Parameter settings.}
    \label{tab:para}
\end{table}

\myparatight{Parameter settings} We use convolutional neural networks as our encoder, decoder, and discriminator. Their architectures are the same as those in HiDDeN and are shown in the Appendix. We set  $\lambda_1=0.7$,  $\lambda_2=0.001$,  batch size $|B|=30$, and watermark length $|w|=30$. We use Adam as optimizer with initial learning rate $\eta$ as $0.001$, and we use CosineAnnealingLR \cite{loshchilov2016sgdr} with default parameter settings to adjust the learning rate. 
For our reinforcement learning based scheduler, we set  $\beta=10$ and  $\mathbf{b}=$ [-0.001, 0.001, 0, 0, 0], where the numbers represent the bias for no post-processing, JPEG, Gaussian blurring, cropping, and resizing, respectively. We set $\alpha=0.2$ and $\gamma=0.5$. We initialize the Q-table as all zeros.  We initialize the exploration probability $\epsilon_0=1.0$ and keep it unchanged in the first 8,000 epochs to explore all possibilities. After that, we decay $\epsilon_t$ by $2.5\times 10^{-4}$ per epoch. We train for 40,000 epochs in total. Table \ref{tab:para} summarizes the parameters.

\begin{table}[!t]
\centering
\subfloat[FaceForensics++]{
\scalebox{1}{
\begin{tabular}{|c|c|c|c|c|c|}
\hline
\multicolumn{1}{|c|}{\multirow{2}{*}{Method}} & \multicolumn{1}{|c|}{\multirow{2}{*}{ACC}} & \multicolumn{1}{c|}{\multirow{2}{*}{FPR}} & \multicolumn{3}{c|}{FNR} \\ \cline{4-6}
                        &  &   & \multicolumn{1}{c|}{FaceSwap1} & \multicolumn{1}{c|}{FaceSwap2} & \multicolumn{1}{c|}{ICface}   \\ \hline
Passive Detector & 66.6 & 0.0 & 0.5 & 100.0 & 100.0 \\ \hline
FaceGuard-ConvWM              &  49.4  & 1.2&  100.0 &100.0           &100.0                     \\ \hline
FaceGuard-HiDDeN              &  86.0    & 1.2&     {0.0}       &   40.5     & 40.2             \\ \hline
FaceGuard           & {99.2}    & 1.7 & {0.0} & {0.0}           & {0.0}                    \\ \hline

\end{tabular}}%
\label{FaceForensics-result}}
\vspace{-1mm}

\subfloat[DFDC]{
\scalebox{1}{
\begin{tabular}{|c|c|c|c|c|c|} 
\hline

\multicolumn{1}{|c|}{\multirow{2}{*}{Method}} & \multicolumn{1}{|c|}{\multirow{2}{*}{ACC}} & \multicolumn{1}{c|}{\multirow{2}{*}{FPR}} & \multicolumn{3}{c|}{FNR} \\ \cline{4-6}
                        &  &   & \multicolumn{1}{c|}{FaceSwap1} & \multicolumn{1}{c|}{FaceSwap2} & \multicolumn{1}{c|}{ICface}   \\ \hline
Passive Detector & 96.1& 0.0 & 0.0 & 21.8 & 1.8 \\ \hline
FaceGuard-ConvWM             & 49.5 & 1.1 & 100.0 &100.0           &100.0                     \\ \hline
FaceGuard-HiDDeN               &   72.2 & 1.3 &   {0.0}     &   72.3         & 90.7             \\ \hline
FaceGuard                 &{99.5}& 1.1 & {0.0} &{0.0}           &{0.0}                    \\ \hline
\end{tabular}}%
\label{DFDC-result}}
\vspace{-2mm}

\subfloat[Trump-Cage]{ 
\scalebox{1}{
\begin{tabular}{|c|c|c|c|c|c|}
\hline
\multicolumn{1}{|c|}{\multirow{2}{*}{Method}} & \multicolumn{1}{|c|}{\multirow{2}{*}{ACC}} & \multicolumn{1}{c|}{\multirow{2}{*}{FPR}} & \multicolumn{3}{c|}{FNR} \\ \cline{4-6}
                        &  &   & \multicolumn{1}{c|}{FaceSwap1} & \multicolumn{1}{c|}{FaceSwap2} & \multicolumn{1}{c|}{ICface}   \\ \hline
Passive Detector & 92.9& 0.0 & 0.7 & 0.0 & 41.7 \\ \hline
FaceGuard-ConvWM               &  49.6 & 0.8 & 100.0 &100.0          &100.0                     \\ \hline
FaceGuard-HiDDeN                  & 74.0  & 1.4 &    10.7     &   83.9      & 57.1             \\ \hline 
FaceGuard                  &{97.7} &   1.5  &   1.0      &   8.1         &{0.0}                     \\ \hline
\end{tabular}}%
\label{Trump-Cage}}
\vspace{-2mm}
\caption{ACC, FPR, and FNR (\%) of different methods.}
\label{tab:overallresults}
\end{table}

\subsection{Experimental Results}

\noindent
{\bf FaceGuard outperforms the compared methods:} Table~\ref{tab:overallresults} shows the results of different methods. We observe that FaceGuard achieves the highest ACC in all the three datasets. Passive detector has high FNRs for deepfakes generated by methods (i.e., FaceSwap2 and ICface) not considered during training in most cases. FaceGuard-ConvWM achieves low ACC because the ConvWM method is not robust against post-processing. In particular, post-processed face images have small SSIMs between the extracted watermarks and the ground truth ones in ConvWM, and thus fake faces are also classified as real when selecting the threshold that achieves at most 1\% FPR on the validation dataset. FaceGuard-HiDDeN achieves smaller ACC because HiDDeN is not robust against JPEG compression. Figure~\ref{bitwiseaccuracy} in Appendix shows that HiDDeN achieves much lower bitwise accuracy than FaceGuard for the extracted watermarks when JPEG compression is used. We found that  HiDDeN and FaceGuard achieve similar bitwise accuracy for the other three post-processing operations.

\begin{table}[!t]
\centering
\begin{tabular}{|c|c|}
\hline
Method & ACC\\ \hline
FaceGuard w.o. JPEG     & 91.2 \\ \hline
FaceGuard w.o. RL       & 98.2 \\ \hline
FaceGuard w.o. RL/JPEG  & 92.4   \\ \hline
FaceGuard           & {99.5}                    \\ \hline
\end{tabular}%
\caption{ACC (\%) of different variants of FaceGuard.}
\label{tab:variants}
\end{table}

\begin{table}[!t]

\parbox{.45\linewidth}{
\centering
\begin{tabular}{|c|c|}
\hline
Dataset & ACC \\ \hline
FaceForensics++   & 99.2 \\ \hline
DFDC              & 93.9\\ \hline
Trump-Cage       & 97.6 \\ \hline
\end{tabular}
\caption{ ACC (\%) of FaceGuard for adaptive deepfakes.}
\label{tab:adaptive}
}
\hfill
\parbox{.45\linewidth}{
\centering
\begin{tabular}{|c|c|}
\hline
Dataset & SSIM \\ \hline
FaceForensics++               &0.94       \\ \hline
DFDC               &0.94       \\ \hline
Trump-Cage               &0.86         \\ \hline
\end{tabular}
\caption{Average SSIM between the real face images and their watermarked versions in FaceGuard for each dataset.}
\label{visualquality}
}
\end{table}

\noindent
{\bf Impact of the two key components of FaceGuard:}  
 FaceGuard has two key components, i.e.,  differentiable JPEG approximation and reinforcement learning based scheduler. We evaluate their impact.  Table~\ref{tab:variants} shows the ACC of different variants of FaceGuard on DFDC dataset. FaceGuard w.o. JPEG uses the original JPEG operation instead of our approximation. As a result, the encoders are not updated for the mini-batches with JPEG compression because the gradients are 0. FaceGuard w.o. RL uses the random scheduler instead of our reinforcement learning based one.  FaceGuard with both components achieves the highest ACC. 

\noindent
{\bf Adaptive deepfakes:} 
An attacker can adapt its deepfakes after  FaceGuard is deployed. The encoder and decoder parameters as well as watermarks in FaceGuard are secret information and not available to an attacker. However, we assume an attacker has access to the architectures of the encoder and decoder as well as the training face images. Therefore, the attacker can train its own encoder and decoder using our training workflow. The attacker extracts a watermark from a watermarked real face image using its own decoder and embeds the corresponding extracted watermark into its fake face image using its own encoder. Table~\ref{tab:adaptive} shows the ACC of FaceGuard against such adaptive deepfakes on the three datasets. We observe that FaceGuard  still achieves high ACC against such adaptive deepfakes. 

\noindent
{\bf Perceptual quality of watermarked face images:} One goal of our watermark is that it should not sacrifice the perceptual quality of the face images. Therefore, we evaluate the perceptual quality of watermarked face images using the widely used SSIM as a metric. Specifically, for each real face image and its watermarked version in FaceGuard, we calculate their SSIM. Table~\ref{visualquality} shows the average SSIM for each dataset. The large average SSIMs  indicate that   the perceptual quality loss introduced by our watermark is small.


\section{Conclusion}
In this work, we propose FaceGuard, a proactive deepfake detection framework.  FaceGuard proactively embeds a watermark into a real face image using a deep-learning-based watermarking method and detects a fake face image if the extracted watermark does not match well with the ground truth one.  Moreover, we design a differentiable approximation of JPEG compression and a reinforcement learning based scheduler to jointly train the encoder and decoder in FaceGuard. Our evaluation results show that FaceGuard can effectively detect deepfakes and outperform existing methods. An interesting direction for future work is to combine passive detection and proactive detection.


\bibliographystyle{plain}
\bibliography{refs}


\section*{Appendix}
\begin{table*}[!h]
\centering
{%
\begin{tabular}{ccccccccc}
\hline
\textit{Layer} & \textit{Type} & \textit{\# Input channel} & \textit{\# Output channel} & \textit{Kernel size} & \textit{Stride} & \textit{Padding} & \textit{With BN} & \textit{Activation} \\
\hline
1              & Conv.        & 3                           & 64                           & (3,3)                & (1,1)           & (1,1)            & True             & ReLU                \\
2              & Conv.        & 64                          & 64                           & (3,3)                & (1,1)           & (1,1)            & True             & ReLU                \\
3              & Conv.        & 64                          & 64                           & (3,3)                & (1,1)           & (1,1)            & True             & ReLU                \\
4              & Conv.        & 64                          & 64                           & (3,3)                & (1,1)           & (1,1)            & True             & ReLU                \\
5              & Conv.        & 97                          & 64                           & (3,3)                & (1,1)           & (1,1)            & True             & ReLU                \\
6              & Conv.        & 64                          & 3                            & (1,1)                & (1,1)           & None             & False            & None  \\

\hline          
\end{tabular}%
}
\caption{Model architecture of the encoder.}
\label{tab:encoder}
\end{table*}

\begin{table*}[!h]
\centering
{%
\begin{tabular}{ccccccccc}
\hline
\textit{Layer} & \textit{Type} & \textit{Input size} & \textit{Output size} & \textit{Kernel size} & \textit{Stride} & \textit{Padding} & \textit{With BN} & \textit{Activation} \\
\hline
1              & Conv.        & 3                           & 64                           & (3,3)                & (1,1)           & (1,1)            & True             & ReLU                \\
2              & Conv.        & 64                          & 64                           & (3,3)                & (1,1)           & (1,1)            & True             & ReLU                \\
3              & Conv.        & 64                          & 64                           & (3,3)                & (1,1)           & (1,1)            & True             & ReLU                \\
4              & Conv.        & 64                          & 64                           & (3,3)                & (1,1)           & (1,1)            & True             & ReLU                \\
5              & Conv.        & 64                          & 64                           & (3,3)                & (1,1)           & (1,1)            & True             & ReLU                \\
6              & Conv.        & 64                          & 64                           & (3,3)                & (1,1)           & (1,1)            & True             & ReLU                \\
7              & Conv.        & 64                          & 64                           & (3,3)                & (1,1)           & (1,1)            & True             & ReLU                \\
8              & Conv.        & 64                          & 30                           & (3,3)                & (1,1)           & (1,1)            & True             & ReLU                \\
9              & GAP        & 30                          & 30                           & --                & --           &  --            & --             & --                \\
10                & Flatten & -- & -- & -- & --& --& --& --\\
11             & FC        & 30                          & 30                           & --                & --           &  --            & False             & None                \\
\hline
\end{tabular}%
}
\caption{Model architecture of the decoder. Input/output size indicates the number of channels for convolution layers and the global average pooling (GAP) layer, while indicating the number of neurons for the fully-connected (FC) layer.}
\label{tab:decoder}
\end{table*}

\begin{table*}[!h]
\centering
{%
\begin{tabular}{ccccccccc}
\hline
\textit{Layer} & \textit{Type} & \textit{Input size} & \textit{Output size} & \textit{Kernel size} & \textit{Stride} & \textit{Padding} & \textit{With BN} & \textit{Activation} \\
\hline
1              & Conv.        & 3                           & 64                           & (3,3)                & (1,1)           & (1,1)            & True             & ReLU                \\
2              & Conv.        & 64                          & 64                           & (3,3)                & (1,1)           & (1,1)            & True             & ReLU                \\
3              & Conv.        & 64                          & 64                           & (3,3)                & (1,1)           & (1,1)            & True             & ReLU                \\
4             & GAP        & 64                          & 64                           & --                & --           &  --            & --             & --                \\
5             & Flatten        & --                          & --                           & --                & --           &  --            & --             & --                \\
6             & FC        & 64                          & 1                           & --                & --           &  --            & False             & None                \\
\hline            
\end{tabular}%
}
\caption{Model architecture of the discriminator. Input/output size indicates the number of channels for convolution layers and the global average pooling (GAP) layer, while indicating the number of neurons for the fully-connected (FC) layer.}
\label{tab:discriminator}
\end{table*}

 \begin{figure*}[!t]
 \subfloat[FaceForensics++]{\includegraphics[width=0.33\textwidth]{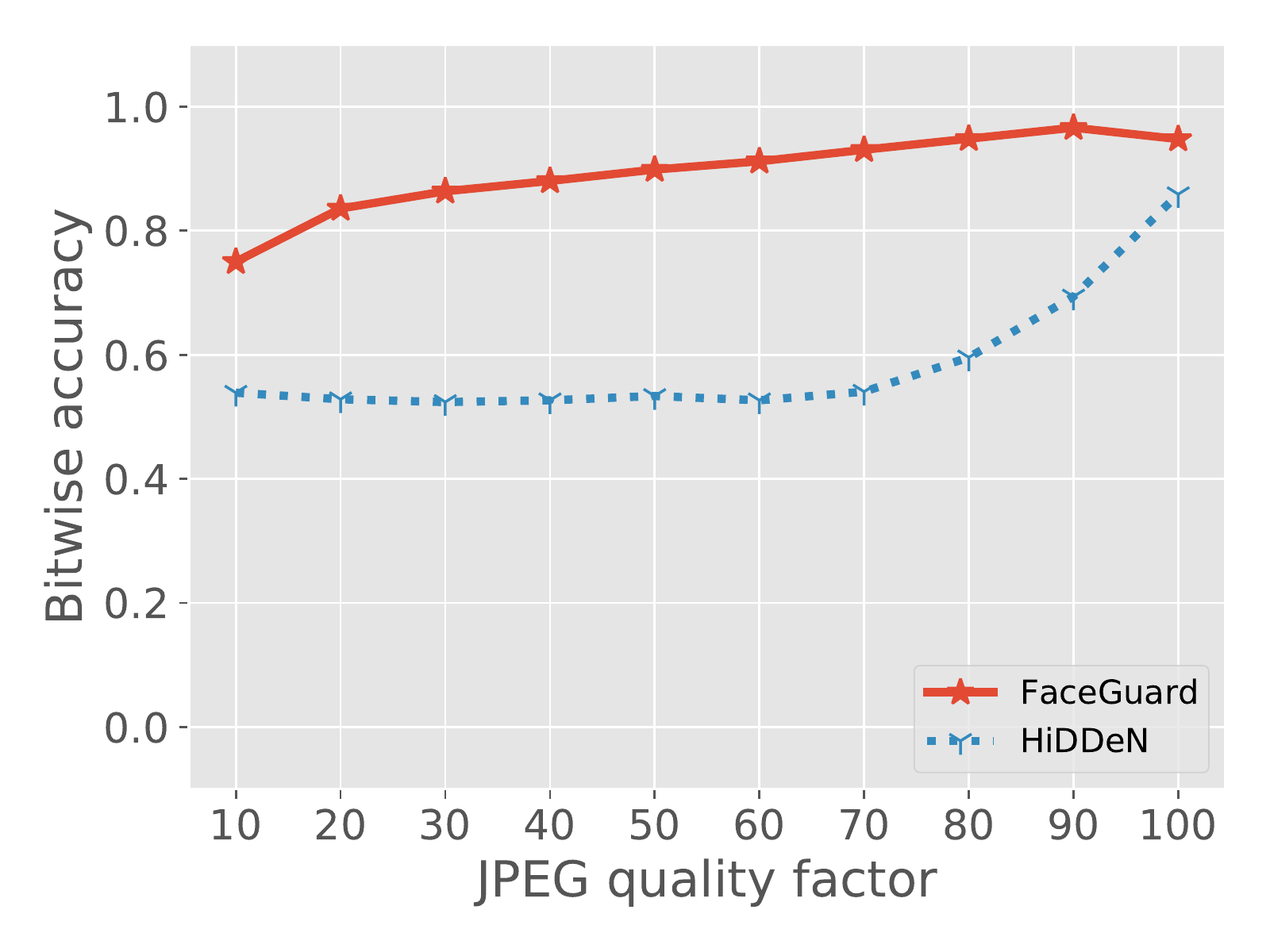}}
 \subfloat[DFDC]{\includegraphics[width=0.33\textwidth]{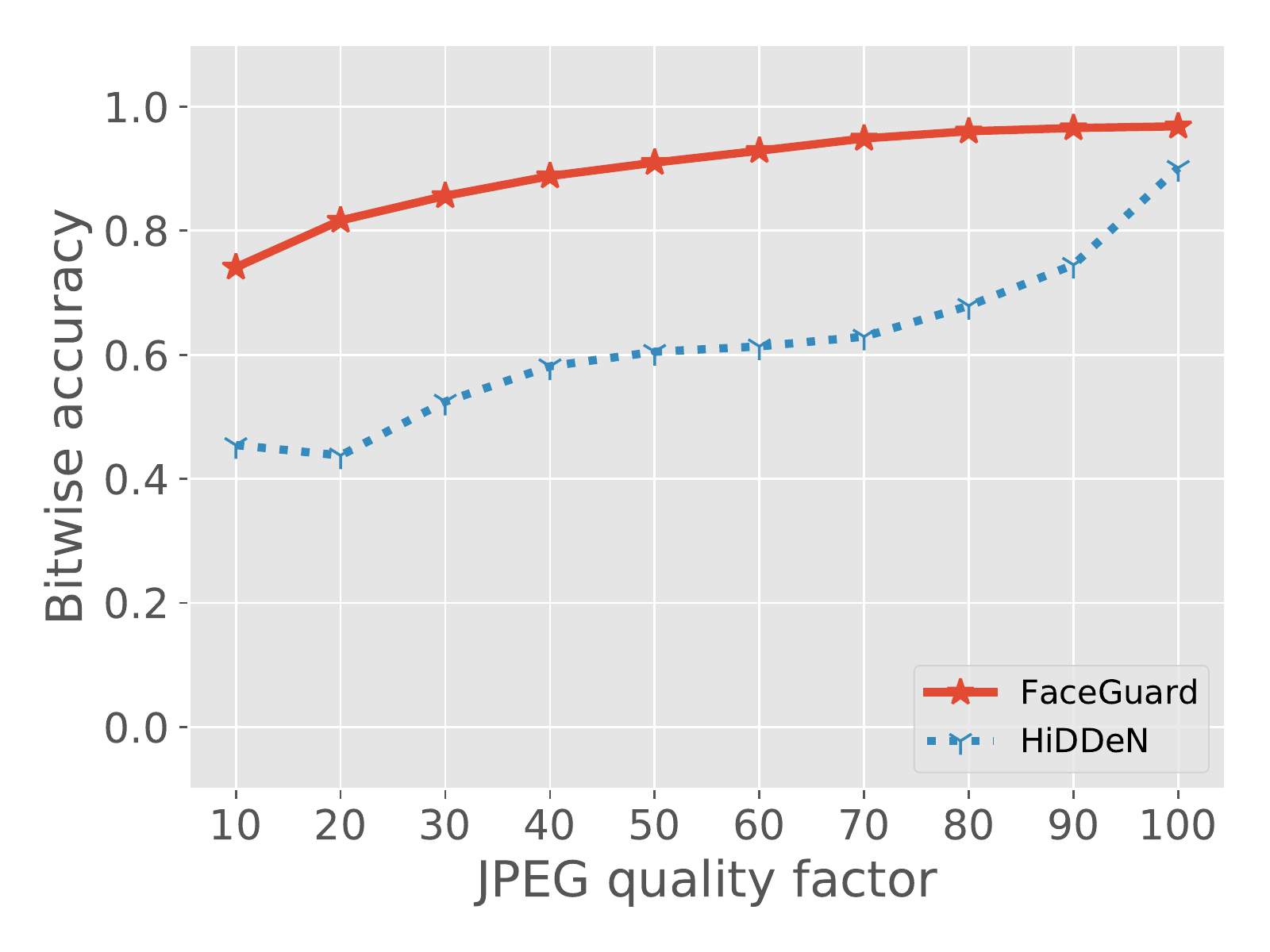}}
 \subfloat[Trump-Cage]{\includegraphics[width=0.33\textwidth]{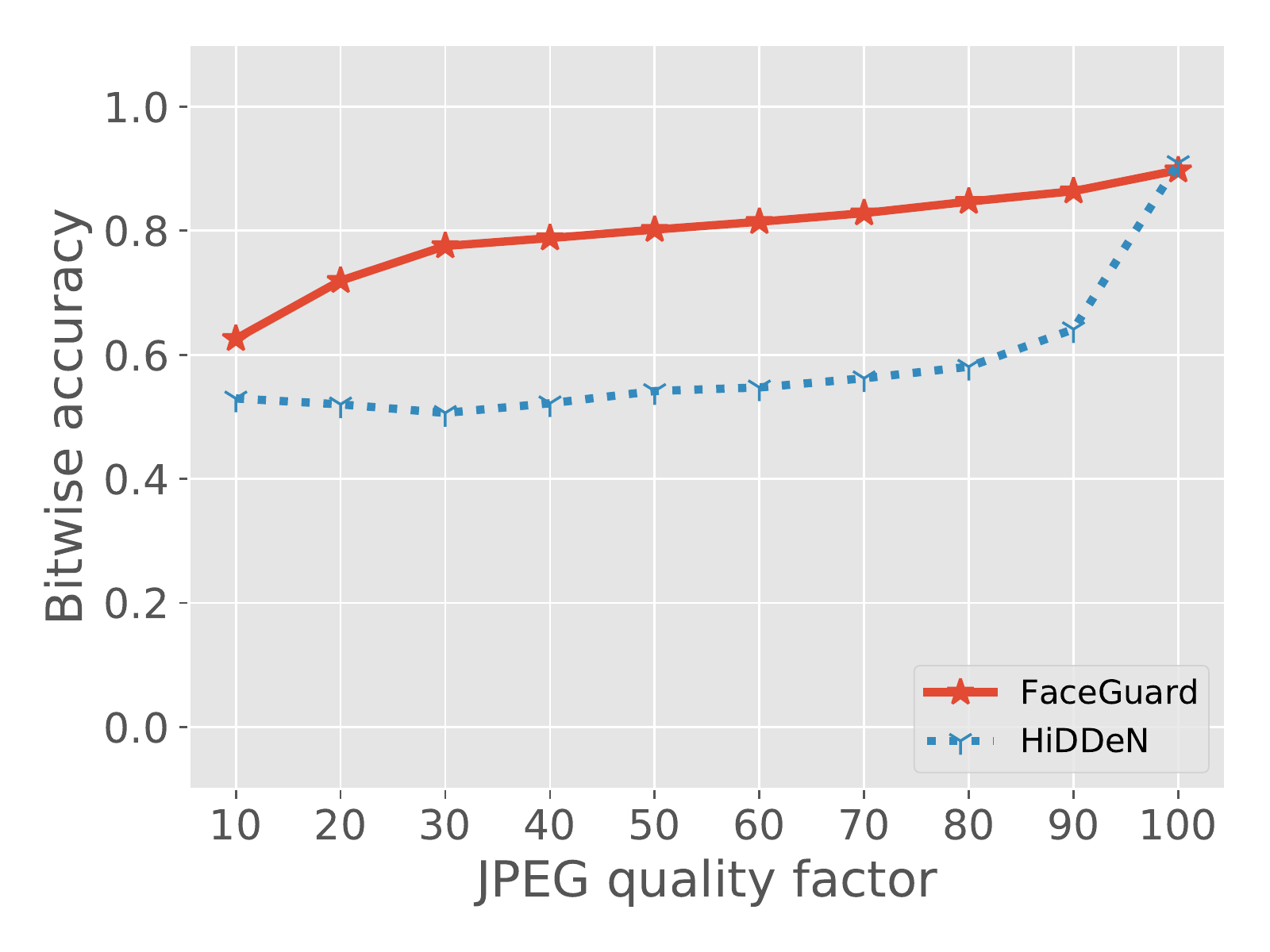}}
\caption{Bitwise accuracy of the extracted watermarks for the real face images post-processed by JPEG compression with different quality factors on the three datasets.}
\label{bitwiseaccuracy}
 \end{figure*}

\end{document}